# A Hierarchical and Attentional Analysis of Argument Structure Constructions in BERT Using Naturalistic Corpora

Liu Kaipeng     Wu Ling
Sichuan International Studies University

**Abstract:** Understanding whether large language models develop human-like representations of abstract grammatical knowledge is a central question in computational linguistics and cognitive science. This study investigates how the Bidirectional Encoder Representations from Transformers (BERT) model processes four fundamental Argument Structure Constructions (ASCs), Resultative, Caused-Motion, Ditransitive, and Way, using a controlled, naturalistic corpus extracted exclusively from fiction genres. We employ a multi-dimensional analytical framework, which integrates MDS, t-SNE as dimensionality reduction, Generalized Discrimination Value (GDV) as cluster separation metrics, Fisher Discriminant Ratio (FDR) as linear diagnostic probing, and attention mechanism analysis. Our results reveal a hierarchical representational structure. Construction-specific information emerges in early layers, forms maximally separable clusters in middle layers, and is maintained through later processing stages. The Way construction is represented as a uniquely distinct schema, underscoring its status as a highly schematic form-meaning pairing. Furthermore, while constructional category is linearly decodable from all major syntactic tokens after layer 2, the model's attention mechanism specializes in diagnosing constructions through the verb-object relation. These findings demonstrate that BERT's internal representations systematically encode constructional abstractions, providing computational evidence for construction grammar theories and establishing a methodology for probing grammatical knowledge in neural language models.

## 1. Introduction

Understanding how the human brain processes and represents the abstract, schematic patterns that underlie language, known as constructions, remains a central challenge in cognitive neuroscience. At the same time, the advent of large-scale artificial neural networks trained on vast text corpora has created powerful new tools for modeling linguistic knowledge. This confluence of fields invites a critical question: do these artificial systems, which achieve remarkable proficiency in language tasks, develop internal representations that mirror the theoretical constructs of human linguistics? This study addresses this question by investigating the processing of Argument Structure Constructions (ASCs) within the Bidirectional Encoder Representations from Transformers (BERT) model, using a carefully controlled dataset of naturalistic language.

Within the framework of Construction Grammar, language is seen as a network of form-meaning pairings that range from specific words to abstract grammatical patterns. ASCs, such as the ditransitive, caused-motion, and resultative constructions, are among the most fundamental of these abstract patterns, governing how verbs integrate with their arguments to form basic clausal meaning. They are therefore prime candidates for probing the grammatical competence of language models. Recent years have seen growing interest in applying constructionist analysis to models like BERT, with studies showing their ability to capture syntactic dependencies and distinguish between some construction types. However, much of this work has relied on synthetically generated or templated sentences, which, while controlled, may lack the variability and ecological validity of naturally occurring language. Furthermore, many studies aggregate data across genres, potentially confounding constructional knowledge with stylistic variation.

To address these limitations, the present study conducts a fine-grained analysis using sentences extracted exclusively from the fiction genre of two major corpora: the British National Corpus and the Corpus of Contemporary American English. We focus on four related yet distinct ASCs: the resultative, caused-motion, ditransitive, and way constructions. By holding genre constant, we aim to isolate the model's handling of constructional syntax and semantics from other sources of variation. Our investigation employs a multi-dimensional analytical framework to ask three specific research questions: (1) How do the internal representations, which is embeddings, for these four ASCs cluster and separate across BERT's twelve processing layers? (2) How is information about c onstruction category linearly encoded in the contextual embeddings of different syntactic roles, like subject, verb, object? (3) Which elements of a sentence does the model's attention mechanism rely upon most to discriminate between these constructions?

To answer these questions, we utilize a suite of complementary techniques, Multidimensional Scaling and t-SNE for visualizing representational geometry; the Generalized Discrimination Value for quantifying cluster separation; linear probe classifiers to assess the accessibility of constructional information; and Fisher Discriminant Ratio analysis of attention weights to identify diagnostically important tokens. This approach allows us to trace the dynamic evolution of constructional representations layer by layer. This research contributes to the interdisciplinary dialogue between computational linguistics and cognitive science. It seeks not only to benchmark the linguistic capabilities of a state-of-the-art AI model but also to use that model as a computational testbed for theoretical linguistic concepts. By demonstrating how, and at which stage of processing, BERT differentiates between core grammatical constructions in natural text, we aim to shed light on both the architecture of a successful language model and the potential computational nature of constructional knowledge itself.

## 2. Literature Review

Since its systematic development by Goldberg in the 1990s (Goldberg, 1995), Construction Grammar has emerged as a pivotal theoretical framework within cognitive linguistics for explaining the nature of linguistic knowledge. This theory posits that the fundamental units of language are pairings of form and meaning-function, known as constructions. These constructions possess independent semantic values that are not fully predictable from their components. Among these, Argument Structure Constructions (ASCs) serve as a class of highly abstract schemas that link verbal semantics to overall sentential meaning, making them essential for understanding language creativity, acquisition, and processing. In parallel, artificial neural network models have achieved transformative success in the field of Natural Language Processing (NLP), making their internal workings a new frontier for understanding intelligence. This convergence of computational linguistics and cognitive linguistics raises a critical question, to what extent do the internal representations of these models, trained on massive corpora, reflect the architecture of linguistic knowledge as described by Construction Grammar? Pursuing the answer not only provides a linguistically rigorous framework for evaluating and interpreting language models but also offers inspiration and validation for computational theories of human language cognition.

In recent years, significant research has focused on dissecting the grammatical knowledge encoded in Large Language Models (LLMs) like BERT and GPT. Foundational work using structural probes demonstrated that BERT's word embedding spaces implicitly encode syntactic tree structures, indicating the model's ability to learn complex grammatical relationships. Subsequent studies further revealed a hierarchical processing profile: shallower layers tend to capture local syntactic and morphological information, while middle-to-deeper layers integrate more complex semantic and discourse-level relations. These findings collectively suggest that LLMs can acquire rich linguistic knowledge through self-supervised learning. However, traditional syntactic frameworks primarily focus on formal rules, often overlooking the holistic form-meaning-function pairings emphasized by Construction Grammar and their usage patterns in specific contexts. Consequently, a new evaluation paradigm is needed to investigate whether models grasp these more holistic and abstract units of linguistic knowledge.

Due to its emphasis on the integration of form and meaning and its network of constructions at varying levels of abstraction, Construction Grammar is increasingly being adopted as a powerful framework for evaluating the deep linguistic capabilities of language models. A series of pioneering studies have demonstrated the potential of this approach. For instance, Li et al. adapted psycholinguistic priming paradigms to Transformer models, providing evidence that LLMs process ASCs as meaningful linguistic units, even activating them in semantically nonsensical "jabberwocky" sentences. Weissweiler et al. explicitly argue that Construction Grammar offers a unique lens for probing neural language models' handling of structural and semantic relationships, uncovering model behaviors that may be missed by purely syntactic or semantic analyses. On the applied front, Sung and Kyle evaluated the performance of pre-trained models like RoBERTa and GPT-4 in identifying ASCs, highlighting their potential utility in linguistic research and educational contexts. Nevertheless, limitations exist in the data and methodologies of these studies. Many rely on template-generated or machine-generated sentences, which, while controlling variables, sacrifice the complexity and authenticity of natural language. Other studies that use corpus data often aggregate materials a cross multiple genres. These risks confounding observed effects with stylistic variation, making it difficult to attribute findings precisely to constructional distinctions.

Despite existing evidence for LLMs' ability to process constructions, several gaps remain. First, regarding data, studies that systematically compare multiple related ASCs using a highly controlled, naturalistic corpus from a single genre are relatively scarce. Second, analytically, most research focuses on final representational outputs or task performance, lacking an integrated and fine-grained examination of how constructional information is dynamically built within the model, how it is distributed across different processing layers, and the specific role played by attention mechanisms. Finally, regarding construction selection, the representation of the distinctive and theoretically important Way construction in computational models remains under-explored. The present study aims to address these gaps. We employ a strictly controlled, naturalistic dataset drawn exclusively from the fiction genre, focusing on four closely related yet distinct ASCs the Resultative, Caused-Motion, Ditransitive, and Way constructions. By employing a multi-dimensional analytical framework, we seek to comprehensively map the dynamic, hierarchical internal representational landscape that the BERT model develops when processing these naturally occurring constructions. This approach allows for a deeper investigation into the fundamental question of how language models acquire and represent abstract grammatical knowledge.

## 3. Methods

This study aims to investigate the representation of four Argument Structure Constructions in the BERT model by strictly controlling the genre of the source corpora. The complete analytical pipeline, encompasses the entire process from corpus collection and preprocessing to multi-faceted model analysis.

### 3.1 Corpus Collection and Construction Annotation

To minimize the potential confounding effect of genre variation on the analysis, the source of the linguistic material was strictly limited to the fiction genre. Fictional language, which blends narrative, descriptive, and dialogic elements, effectively represents natural, coherent written language use and provides an ideal context for studying grammatical constructions.

### 3.1.1 Construction Definitions and Retrieval Strategy

Four ASCs that are distinct in their syntactic and semantic properties were selected.

(1) Resultative Construction: Signifies an action causing a change of state in a patient, like "She painted the wall red".

(2) Caused-Motion Construction: Signifies an agent causing the motion of a theme along a path, like "He pushed the cart into the garage".

(3) Ditransitive Construction: Signifies an agent transferring a theme to a recipient, like "She gave him a book".

(4) Way Construction: Signifies an agent creating a path or making progress by means of the action denoted by the verb, like "He fought his way to the top".

To efficiently and accurately extract instances of these constructions from large corpora, instances of these constructions were systematically extracted from the fiction sub-corpora of the British National Corpus (BNC) and the Corpus of Contemporary American English (COCA) using part-of-speech (POS) based query patterns, illustrated in table 1. To ensure data quality and balance, 100 valid instances of each construction were randomly sampled from each corpus, resulting in a balanced dataset of 800 sentences total, 200 sentences per construction. All sentences were manually verified to confirm their compliance with the target construction's definition.

| Construction Type | BNC Query (C5 Tagset) | Interpretation | COCA Query | Interpretation |
|---|---|---|---|---|
| Resultative | _VV* (* _NN*\|_PN*) (_AJ*) | A verb + (a noun or pronoun) + an adjective | VERB * NOUN ADJ | A verb + any word(s) + a noun + an adjective |
| Caused-Motion | _VV* * _NN* _AVP * * *_NN* | A verb + any word(s) + a noun + an adverb phrase + (complex path phrase) + a noun | VERB * NOUN PREP | A verb + any word(s) + a noun + a preposition |
| Ditransitive | _VV* | A verb + (a | VERB PRON * | A verb + a |

| | | | | | | | |
|---|---|---|---|---|---|---|---|
| | (_PN*\|_NP0) *_NN* | | pronoun or proper noun) + any word(s) + a noun | NOUN | | personal pronoun + any word(s) + a noun | |
| Way | _VV* way | _DPS | A verb + a possessive determiner + "way" | VERB way | POSS | A verb + a possessive pronoun + "way" | |

Table 1 the query patterns in BNC and COCA

Note: * matches any number (≥0) of words; | means "or"; _VV* matches all verb forms; _AJ* matches all adjective forms; _AVP matches adverb phrases.

The retrieval expressions employed in this study were designed to strike a balance between efficiency in corpus querying and accuracy in instance retrieval. These patterned expressions maximize the recall of potential target construction instances, which are subsequently filtered through an essential manual screening step to remove sentences that do not meet strict definitions. This approach avoids the significant upfront labor associated with crafting overly complex and restrictive queries, while still ensuring the purity and reliability of the final dataset.

### 3.2 Token Alignment and BERT Encoding

To enable a meaningful comparative analysis across sentences with differing surface structures, a robust pipeline for syntactic annotation and token alignment was implemented prior to model inference.

#### 3.2.1 Syntactic Role Annotation

Each of the 800 sentences was first processed using SpaCy (en_core_web_sm model) to perform part-of-speech tagging and dependency parsing. The output was manually reviewed to correct any parsing errors and to consistently annotate the core syntactic roles relevant to the ASCs under study. For each sentence, we identified and labeled:

Subject (SUBJ): The nominal head of the subject noun phrase

Verb (VERB): The main lexical verb of the construction.

Direct Object (OBJ): The nominal head of the primary object noun phrase like "the wall" in the resultative, "the cart" in the caused-motion.

Indirect Object (OBJ2): For ditransitive sentences, the recipient noun phrase, like "him".

Preposition (PREP): For caused-motion sentences, the head of the directional prepositional phrase, like "into".

Keyword "way": For way construction sentences, the noun "way" was specifically tagged.

#### 3.2.2 BERT Tokenization and Role-to-Token Mapping

Sentences were then tokenized using the WordPiece tokenizer of the bert-base-uncased model. A critical step involved mapping the annotated syntactic roles (e.g., SUBJ: "0") to the corresponding sequence of BERT sub-word tokens (e.g., ["artist"] or, for an unknown word, ["art", "##ist"]). The mapping rule was as follows: for any lexical item split into multiple sub-word tokens, the embedding of its first sub-word token was selected as the representative vector for the entire word. This approach ensured that each syntactic role corresponded to a single, consistent 768-dimensional vector per layer for analysis

#### 3.2.3 Embedding and Attention Weight Extraction

The aligned token sequences were fed into the pre-trained BERT model without any fine-tuning. For each sentence, we extracted two types of data from all 12 Transformer encoder layers. First is Contextualized Embeddings, which is the 768-dimensional output vector for every token in the sequence. The second is Attention Weight Matrices, the 12 x (Sequence Length) x (Sequence Length) matrices (one per attention head) detailing the attention paid between every pair of tokens. This process resulted in a dataset where, for each sentence, we had access to the evolving representations and attention patterns of its core syntactic components across BERT's processing hierarchy

### 3.3 Analytical Framework

Our investigation employed a multi-faceted analytical framework designed to probe different aspects of BERT's internal representations concerning the four ASCs

#### 3.3.1 Dimensionality Reduction for Cluster Visualization

To gain an intuitive understanding of how sentences group by construction type in high-dimensional space, we applied two dimensionality reduction techniques to the embeddings of the CLS token at each layer.

In Multidimensional Scaling, we used classic MDS (sklearn.manifold.MDS) with a Euclidea n distance metric to project the 768-D CLS embeddings onto a 2D plane, preserving the global pairwise distances as faithfully as possible. This provided a stable, parameter-free overview of cluster structure.

In t-Distributed Stochastic Neighbor Embedding, we employed t-SNE (sklearn.manifold.TSNE, perplexity=30) to create a complementary visualization that emphasizes local similarities and may reveal finer sub-structures within clusters. Points in the resulting 2D plots were color-coded by their construction label, Resultative, Caused-Motion, Ditransitive, Way.

### 3.3.2 Quantifying Cluster Separation: Generalized Discrimination Value

To move beyond qualitative visualization and obtain a robust, scalar measure of how well the embeddings separate by construction type, we calculated the Generalized Discrimination Value for the CLS token embeddings at each layer, 0 through 12, where layer 0 refers to the initial input embedding. The GDV is calculated based on the difference between the mean intra-class distance and the mean inter-class distance, normalized by dimensionality. A more negative GDV indicates better cluster separation, 0 indicates chance-level overlap. This provided a layer-by-layer profile of construction-specific information concentration.

### 3.3.3 Probing for Linearly Accessible Information

To test whether the construction category is encoded in a linearly decodable manner within the embeddings of specific syntactic roles, we conducted linear probing experiments. For each layer *l* and for each key token type (CLS, VERB, OBJ), we trained a simple 4-class Linear Support Vector Machine (SVM) classifier (sklearn.svm.LinearSVC) using the corresponding 768-dimensional token embeddings as features and the construction label as the target. Classifiers were evaluated using 5-fold cross-validation, and the mean classification accuracy was reported. Accuracy significantly above the random chance baseline of 25% indicates that information useful for distinguishing the four constructions is present and readily accessible in that specific token's representation at that layer

### 3.3.4 Analyzing Discriminative Attention Patterns

To understand how BERT's attention mechanism contributes to differentiating between constructions, we analyzed the attention weights using the FDR. For each attention head in each layer, we first derived a feature for each key token, like OBJ, the sum of attention weights received by that token from all other tokens in the sentence. This value represents the total "incoming attention" for that syntactic role. We then treated the set of these values for one token across all sentences as a distribution and calculated its pairwise FDR between every two construction types. The formula for FDR between class *a* and *b* is:

$$FDR = \frac{(Mean\_a - Mean\_b)^2}{(Variance\_a + Variance\_b)}$$

A high FDR for a specific token, like OBJ, in a specific attention head indicates that the pattern of attention allocated to that token is highly distinctive for that pair of constructions. By averaging FDRs across all construction pairs and identifying consistent patterns across layers and heads, we could determine which syntactic roles were most discriminative within the attention mechanism.

### 4.1 Dimensionality Reduction Visualization of Construction Clusters

Figure 1 the MDS in each layer

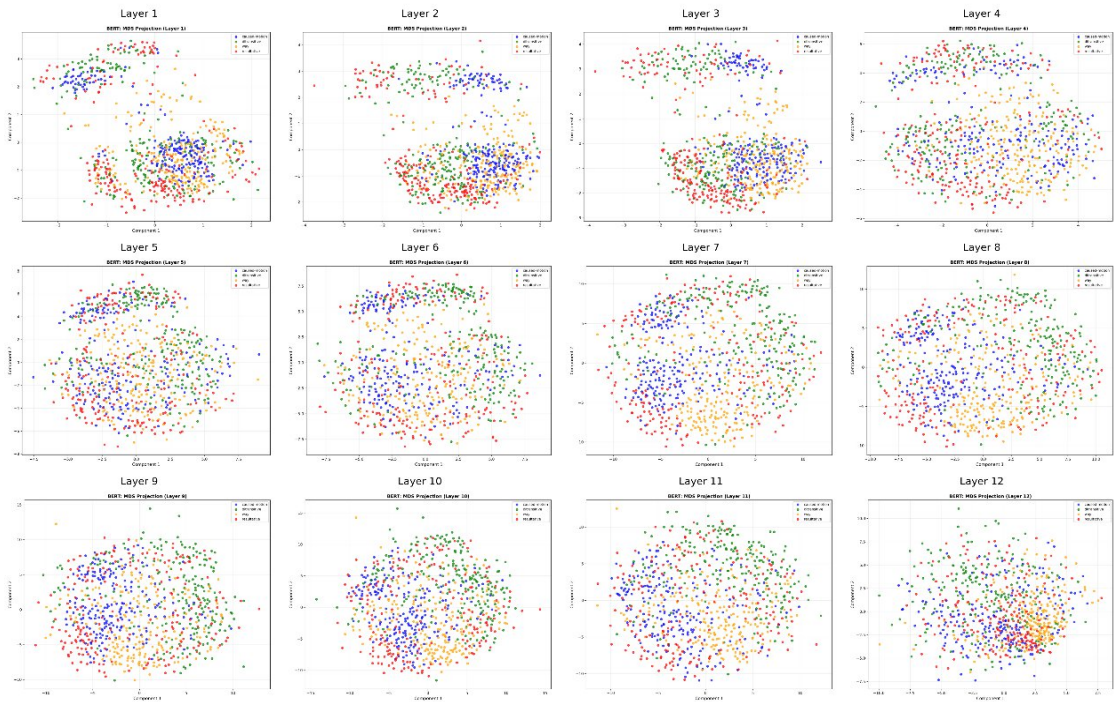

To visualize how BERT internally organizes sentences from the four distinct Argument Structure Constructions (ASCs), we employed Multidimensional Scaling (MDS) on the contextualized embeddings of the [CLS] token across all 12 layers. The [CLS] token is designed to aggregate sentence-level information, making it a suitable probe for overall construction representation. The MDS projections, as illustrated in Figure 1, reveal a clear and dynamic evolution of representational geometry across BERT's processing depth. The narrative can be segmented into three key phases. Undifferentiated Soup in Early Layers (1-2). In the first and second layers, the [CLS] embeddings for all four constructions, Resultative, Caused-Motion, Ditransitive, and Way, show extensive overlap within a centralized cloud. This indicates that the initial contextualization provided by these layers does not yet segregate sentences based on their overarching argument structure schema. Representations are likely dominated by shallow lexical and local syntactic features.

Emergence of Construction-Specific Clusters in Middle Layers (3-6). A striking transformation occurs beginning in layer 3. The point cloud separates into discernible, distinct clusters corresponding to each of the four ASCs. This separation becomes most pronounced in layers 4 and 5. Notably, the Way construction forms the most compact and isolated cluster, suggesting its unique syntax-semantics (e.g., the obligatory and non-literal "way" object) is highly distinct and consistently encoded. The Resultative and Caused-Motion constructions, while separable, are often positioned in relative proximity, reflecting their shared semantic core of "causality." The Ditransitive construction occupies its own distinct region in the space.

Refinement and Potential Functional Specialization in Later Layers (7-12). In the deeper layers, the clear cluster boundaries observed in the middle layers persist but undergo subtle refinement. Clusters may show slight dispersion or shifts in relative positioning. This phase is consistent with the hypothesis that later layers in Transformer models specialize in integrating information for downstream task-specific predictions, potentially modulating the pure constructional representations formed earlier.

This layered trajectory aligns with and extends findings from prior work on LSTM networks and BERT with synthetic data (Ramezani et al., 2024). It demonstrates that BERT, when processing naturalistic language, progressively builds hierarchical representations of grammatical constructions. The formation of maximally separated clusters in the middle layers suggests these layers are crucial for abstracting and distinguishing core clausal patterns, before further contextual and pragmatic fine-tuning occurs in the final layers. The distinctiveness of the Way construction cluster provides compelling computational evidence for its status as a highly schematic and uniquely identifia ble form-meaning pairing in English grammar.

**4.2 Quantitative and Diagnostic Analyses of Construction Representation**

Beyond visualization, we employed quantitative metrics and diagnostic classifiers to precisely measure the separability and accessibility of construction-specific information across BERT's layers.

**4.2.1 Generalized Discrimination Value (GDV) Analysis**

Figure 2 the GDV across layers

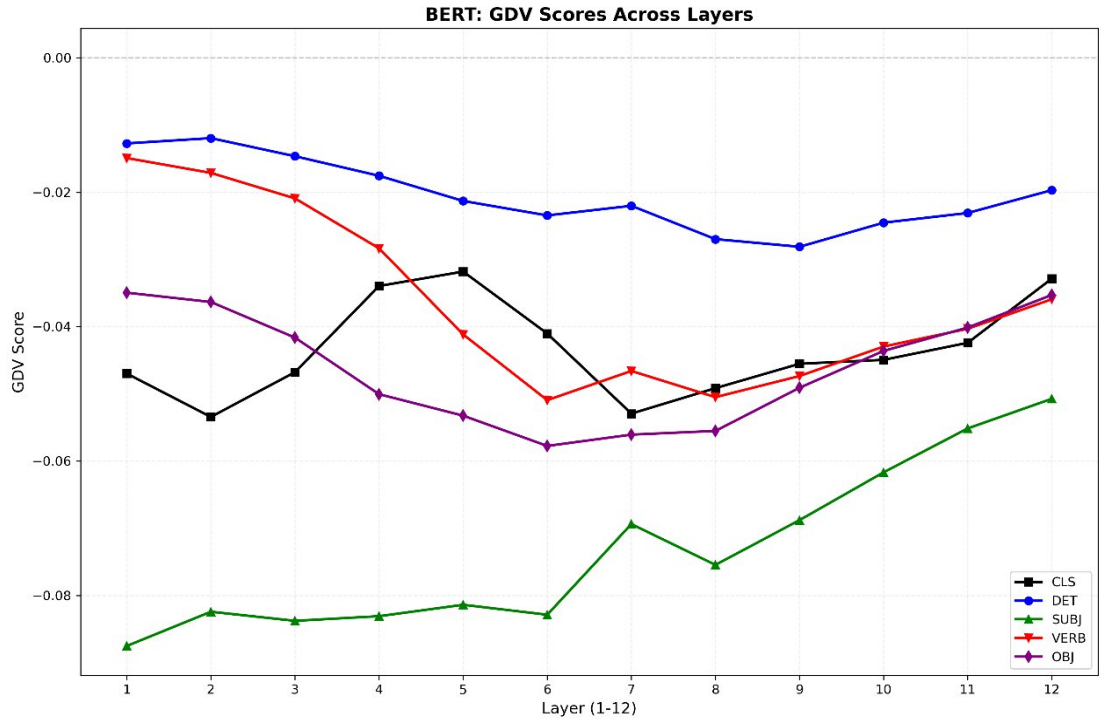

To move beyond qualitative visual assessment, we computed the Generalized Discrimination Value (GDV) for the [CLS] token embeddings at each layer. The GDV quantifies cluster separation, with more negative values indicating better discrimination between construction types.

As shown in Figure 2, the GDV profile provides a precise, scalar confirmation of the pattern observed in the MDS plots. The GDV decreases sharply from Layer 1, reaching its global minimum (approximately -0.072) at Layer 4. This identifies Layer 4 as the point of optimal linear separability between the four ASCs based on the aggregate sentence representation. Following this peak, the GDV becomes slightly less negative in subsequent layers, plateauing in the deeper half of the network. This trajectory reinforces the interpretation that constructional schemas are most explicitly and distinctly represented in the middle layers of BERT, after which representations may be refined or transformed for other purposes.

**4.2.2 Linear Probing of Token Embeddings**

We investigated whether construction category information is not only present but also linearly accessible in the contextual embeddings of specific syntactic roles. To this end, we trained simple linear classifiers (probes) to predict the construction type from individual token embeddings at each layer. The results, detailed in Figure 3, reveal a striking pattern.

Figure 3 the probe classification accuracy across layers

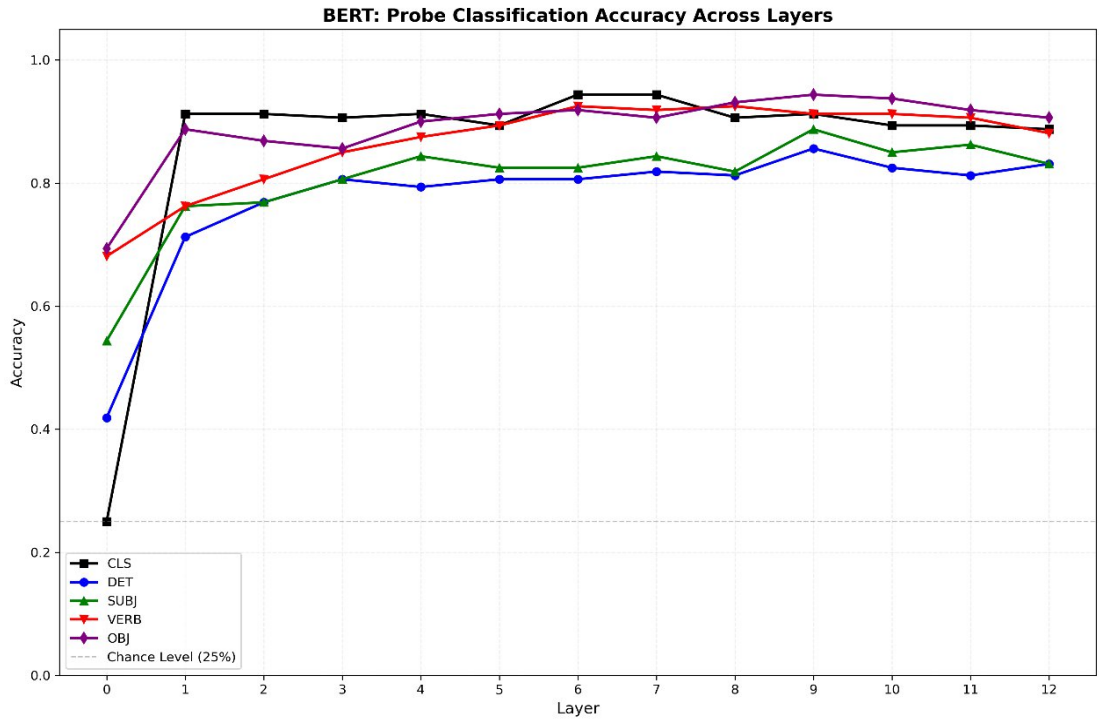

Layer 1 Baseline: All probes perform at or near chance level (25%), confirming that initial contextual embeddings contain minimal explicit information about the high-level construction type. Sharp Rise in Middle Layers: A dramatic accuracy jump occurs at Layer 2, where all token probes (CLS, SUBJ, VERB, OBJ, DET) achieve classification accuracy exceeding 90%. Sustained High Accessibility: This high level of accuracy (>90%) is maintained robustly across all subsequent layers (3-12) for all probed tokens. Notably, the probes for the VERB and CLS tokens consistently show the highest accuracy, often above 95%.

This finding demonstrates that by Layer 2, discriminative features for classifying ASCs become linearly encoded in the contextual representation of every major syntactic constituent. The sustained high accuracy indicates that this information remains readily decodable throughout the network's depth, forming a stable latent representation of constructional identity.

**4.2.3 Attention-Based Discrimination via Fisher Discriminant Ratio**

To understand which elements of the sentence BERT's attention mechanism "attends to" when differentiating constructions, we analyzed the attention weights using the FDR. The FDR measures how distinct the attention pattern directed at a particular token is across different construction types.

Figure 4 the FDR scores for each layer

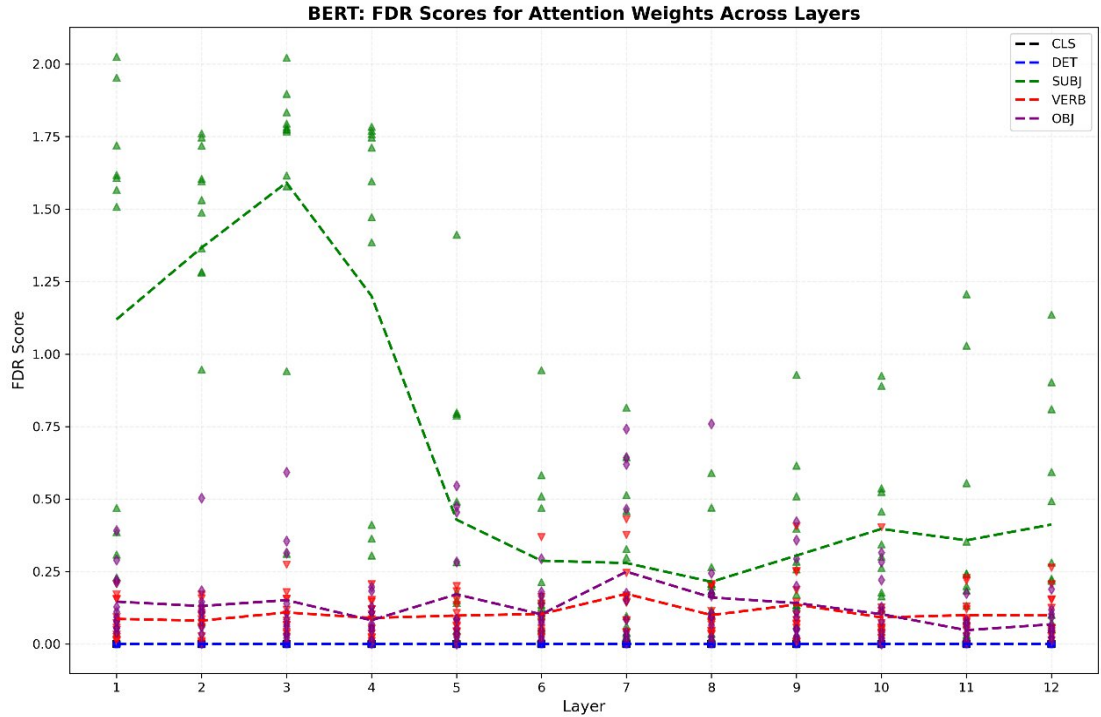

Figure 4 presents the mean FDR scores for key tokens across layers and attention heads. The results highlight a clear hierarchy of diagnostic importance within the attention mechanism. The Object (OBJ) has the primary role. The OBJ token exhibits the highest FDR scores across almost all layers, with a prominent peak in middle layers (4-7). This indicates that how much attention is paid to the direct object, and from where, is the single most distinctive signal for distinguishing between the four ASCs. The Verb (VERB) has secondary role. The VERB token shows the second-highest discriminative power, following a similar layer-wise trend as the OBJ. This underscores the verb's central role in defining the argument structure frame. The Subject and CLS has minimal role. In contrast, the SUBJ and CLS tokens show consistently low FDR scores. This suggests that while their embeddings contain construction information, as shown by probing, the attention patterns directed at them are not primary sources of discrimination for these particular constructions.

This analysis reveals a dissociation: whereas construction information is linearly accessible from all token embeddings after Layer 2 (per probing), the attention mechanism itself is specialized to focus on and extract diagnostic cues primarily from the core VERB-OBJ relationship. This finding aligns with linguistic intuition that the verb and its relationship to its primary object are fundamental in defining an argument structure construction.

**4.3 Refined Cluster Analysis via t-Distributed Stochastic Neighbor Embedding**

To complement the global perspective offered by MDS and to examine the local structure within the construction clusters in greater detail, we employed t-Distributed Stochastic Neighbor Embedding (t-SNE). While MDS preserves global pairwise distances, t-SNE prioritizes the preservation of local neighborhoods, making it particularly sensitive to the emergence of fine-grained substructures.

Figure 5 t-SNE projections by layer

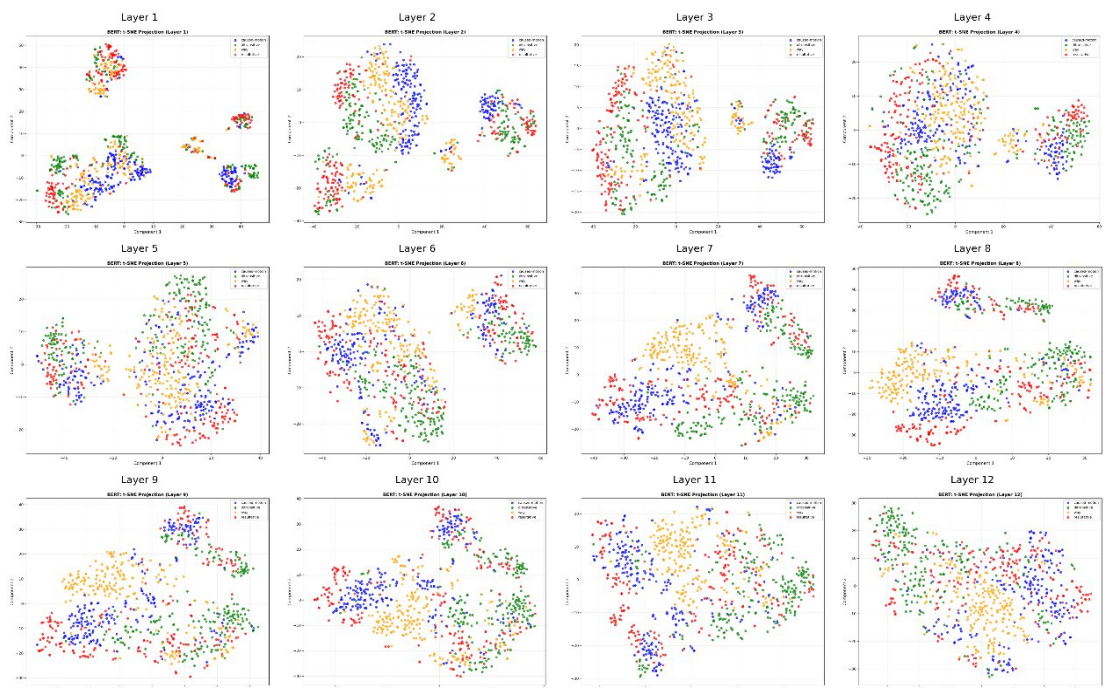

The t-SNE projections (Figure 5) largely corroborate the overall developmental narrative captured by MDS: a progression from overlap in early layers to separated clusters in middle layers, followed by stable but slightly modulated representations in later layers. However, t-SNE provides several critical, nuanced insights that enrich our understanding. First is the validation of core clusters. The clear separation of the four main construction types is unequivocally confirmed. The Way construction again appears as a uniquely dense and isolated cluster, reinforcing its distinct status. The second is revelation of intrinsic substructures. More importantly, t-SNE visualizations frequently reveal prominent substructures within the main construction clusters, particularly for the Ditransitive and Way constructions. For example, the Ditransitive cluster in layers 4-6 may split into two or more discernible sub-clusters. This suggests that BERT's representation internally differentiates between sub-variants of the same abstract schema, potentially based on semantic verb classes or other lexico-syntactic features.

The third is Highlighting Local Similarities. The technique sometimes shows tighter local grouping between sentences from different constructions that share strong lexical or thematic similarities, which are less apparent in the MDS plots. This underscores that while high-level constructional categories are robustly represented, the model's geometry also preserves finer-grained similarity relationships.

The substructures revealed by t-SNE indicate that BERT's internal representations capture not only the abstract, schematic level of argument structure constructions but also more granular levels of linguistic organization. The clustering of sentences within a construction type implies that the model learns a family-resemblance structure, where instances are organized by shared lower-level features even as they instantiate the same high-level pattern. This finding aligns with usage-based constructionist theories, where constructions exist at varying levels of semanticity and are linked in a hierarchical network. The t-SNE analysis thus provides evidence that BERT's representational space mirrors this property of the mental construct-i-con.

## 5. Discussion

The present study systematically investigated the representation of four Argument Structure Constructions in the BERT language model using a controlled, naturalistic corpus drawn exclusively from fiction. By employing a multi-method analytical framework, encompassing dimensionality reduction, cluster separation metrics, linear probing, and attention mechanism analysis. We have delineated a coherent and detailed picture of how this transformer-based architecture processes and distinguishes between these fundamental grammatical schemas. Our findings both converge with a nd extend prior research, offering new insights into the computational nature of constructional knowledge.

A central finding of this study is the layered trajectory of construction representation. The progression from undifferentiated embeddings in early layers (1-2) to maximally separated clusters in middle layers (3-6), followed by stabilized or slightly modulated representations in later layers (7-12), reveals a process of progressive abstraction and specialization. This aligns with the "BERT Rediscovers the NLP Pipeline" hypothesis (Tenney et al., 2019a), where middle layers have been shown to capture syntactic and semantic role information. Our work specifies that this includes the abstraction of construction-level schemas. The peak of cluster separation (GDV minimum) and attentional discriminability (FDR peaks for OBJ/VERB) in layers 4-7 suggests these are crucial constructional layers where the model integrates verb semantics, argument roles, and syntactic frames into a unified, category-distinctive representation before further contextual integration.

The consistent and pronounced isolation of the Way construction across all analytical methods provides compelling computational evidence for its theoretical status. Its compact cluster in visualizations and the high discriminative power of its obligatory "way" object in attention analysis underscore that it is a highly schematic and uniquely identifiable form-meaning pairing. This finding resonates with construction grammar's emphasis on unique or extra syntax (Goldberg, 1995). Our results demonstrate that BERT is sensitive to such diagnostics, not merely to verb-centered argument structure. The high probing accuracy for all tokens from layer 2 onward further confirms that constructional information is a pervasive property of the contextualized sentence representation, not localized to a single component.

An intriguing dissociation emerged between the results of linear probing and attention (FDR) analysis. While construction category information was linearly accessible from the embeddings of all major syntactic tokens (SUBJ, VERB, OBJ) after layer 2, the attentional discriminability was overwhelmingly focused on the OBJ and VERB tokens. This suggests that BERT solves the task of distinguishing ASCs by specializing its attention mechanism to scrutinize the core predicate-argument nexus (the verb and its primary object), which is linguistically the most informative region for determining clause structure. The subject and functional elements, while their embeddings encode the constructional context, do not require differential attention patterns for this discrimination task. This offers a mechanistic insight: the model builds a distributed, linearly decodable representation of the construction across the sentence, but achieves this through a focused attention strategy on key relational elements.

This study validates the utility of construction grammar as a probing framework for neural language models (Weissweiler et al., 2023a). By using closely related ASCs from a single genre, we mitigated lexical and stylistic confounds, allowing us to more confidently attribute observed patterns to constructional differences. The success of simple linear probes in decoding construction categories from intermediate layers suggests that constructional schemas emerge as linearly separable directions in BERT's representational space. This has implications for interpretability, suggesting that concepts from linguistic theory can sometimes correspond to understandable geometries in model embeddings.

Several limitations of the current study pave the way for future research. First, while the focus on fiction controls for genre, it limits the generalizability of findings. Future work should test whether the identified constructional layers and the representational hierarchy are stable across genres like academic text or spoken dialogue. Second, we analyzed the static representations of individual sentences. Investigating how these constructional representations are dynamically built and used during sentence processing would bridge computational findings with psycholinguistic paradigms. Third, extending this methodology to a broader spectrum of constructions, including more abstract and idiomatic ones, would further test the limits and capabilities of LLMs in capturing the full richness of the construction. Finally, the logical next step, as indicated in the introduction of the original paper, is to correlate these computational findings with neuroimaging data to search for neural correlates of such hierarchical construction processing.

Our findings not only illuminate how BERT encodes argument structure constructions (ASCs), but also contribute to a deeper understanding of how different neural architectures represent abstract linguistic knowledge. Crucially, by analyzing naturally occurring language from large corpora (BNC/COCA), our study offers high ecological validity, revealing how pre-trained language models internalize constructions as they appear in real human usage, rather than in artificially controlled settings.

A key insight from our analysis is the layer-wise emergence of constructional abstraction. Using geometric distance variance (GDV), probing classifiers, and dimensionality reduction (MDS/t-SNE), we consistently observe that BERT achieves maximal separation among the four ASCs, not at the final layer, but around Layer 4. This suggests that mid-level layers in BERT specialize in encoding abstract, construction-level schemas, while higher layers may shift toward task-specific or discourse-oriented processing. This pattern aligns with the pipeline hypothesis of contextualized representations (Tenney et al., 2019a), wherein syntactic and semantic abstractions crystallize before pragmatic or referential details are resolved.

This observation gains further significance when contrasted with prior work on recurrent architectures. Ramezani et al. (2025), in a parallel study using synthetically generated data, report that LSTMs exhibit peak construction sensitivity only in their final hidden layer, a consequence of their sequential, incremental processing. In contrast, BERT's early-to-mid layer specialization reflects the power of its self-attention mechanism, which enables simultaneous integration of long-range dependencies across the entire input sequence. Thus, the Transformer architecture appears uniquely suited to rapidly construct global, schematic representations, precisely the kind of form-meaning pairings posited by construction grammar (Goldberg, 1995, 2006).

Moreover, our results provide compelling computational support for the psychological reality of constructions. The fact that BERT, trained solely on next-token prediction, spontaneously organizes its internal representations into clusters that mirror theoretically defined ASC categories suggests that these constructions are not merely analytical conveniences, but robust, learnable patterns in language data. Notably, the Way construction emerges as a highly compact and isolated cluster, consistent with its status as a highly schematic, idiosyncratic construction in linguistic theory. This convergence between model behavior and linguistic intuition strengthens the case for treating constructions as fundamental units of linguistic cognition.

Importantly, our fine-grained attention analysis reveals a critical dissociation: high information content does not necessarily correlate with high attention weight. For instance, function words like the or to, though carrying low semantic load, often receive substantial attention, while key lexical verbs may be relatively down weighted. This challenges simplistic interpretations of attention as importance and instead supports a view of attention as a mechanism for structural scaffolding, orchestrating syntactic roles and dependency relations rather than highlighting semantic content per se. This nuance deepens our understanding of how BERT builds compositional meaning.

From a cognitive science perspective, our work positions large language models as computational probes of human language processing. If constructions are indeed part of the human mental lexicon, then a model that successfully acquires them from exposure alone may serve as a testbed for generating neuroscientific hypotheses. Future fMRI or MEG studies could investigate whether human brains show analogous layer-like processing stages, like, mid-level cortical regions encoding constructional schemas, mirroring BERT's representational geometry. Such cross-disciplinary validation would lend strong support to the neural reality of constructions.

Finally, our t-SNE visualizations reveal not only clear separation between constructions but also internal substructure within categories. For example, distinct clusters within the Ditransitive construction corresponding to verb classes like give-type vs. tell-type. This suggests that BERT captures not just abstract schemata, but also the gradient, usage-based variation that defines construction networks in usage-based theories (Bybee, 2010). Thus, our findings bridge formal and functional approaches, showing how statistical learning over natural language can give rise to both categorical and probabilistic aspects of grammatical knowledge.

In sum, by combining corpus-driven realism with multi-faceted representational analysis, this study tells a richer story: one in which BERT's internal dynamics not only reflect the architecture of modern AI systems, but also resonate with long-standing theories about how humans store, process, and generalize linguistic knowledge. Far from being a black box, BERT emerges here as a powerful lens through which we can re-examine foundational questions in linguistics and cognitive science.

## 6. Conclusion

Using a controlled corpus of naturalistic fiction, this study provides a systematic, multi-faceted analysis of how the transformer-based language model BERT represents four fundamental Argument Structure Constructions. By integrating geometric visualization, quantitative clustering metrics, linear diagnostic probes, and attention mechanism analysis, we have mapped the hierarchical process through which abstract grammatical schemas emerge and are maintained within the model's architecture.

Our key findings converge to paint a coherent picture: BERT constructs a layered representation of grammatical knowledge, where early layers capture local features, middle layers, notably layers 3-7 specialize in forming highly separable, construction-specific clusters, and later layers refine these representations for broader contextual integration. The distinct and isolated encoding of the Way construction underscores the model's sensitivity to unique, schematic form-meaning pairs. Furthermore, we identified a revealing dissociation: while constructional identity is linearly decodable from the embeddings of all core syntactic tokens after the second layer, the model's attention mechanism is strategically optimized to focus discriminative power on the central verb-object relation, highlighting it as the core diagnostic for argument structure.

These results have a dual significance. For linguistic theory, they offer compelling computational evidence supporting constructionist approaches, demonstrating that abstract, holistic clausal patterns emerge as stable, linearly separable regions in a model trained solely on form. For the field of AI and model interpretability, this work illustrates that modern LLMs develop internal representations that mirror theoretically relevant linguistic abstractions. The methodology establishes construction grammar as a powerful framework for probing model understanding, revealing that the black box encodes grammatical knowledge in a structured, hierarchical, and partially interpretable manner.

Ultimately, this research strengthens the bridge between computational modeling and the cognitive science of language. It suggests that the learning pressures of next-word prediction on vast natural data are sufficient to drive a system towards building human-like structural abstractions, offering a new lens through which to explore the timeless question of how form and meaning are linked in the mind.